\documentclass[journal]{IEEEtran}
\IEEEoverridecommandlockouts

\usepackage{cite}
\usepackage{graphics} 
\usepackage{epsfig} 
\usepackage{mathptmx} 
\usepackage{times} 
\usepackage{amsmath} 
\usepackage{amssymb}  
\usepackage{multirow}
\usepackage{booktabs}
\usepackage{fourier-orns}
\usepackage{setspace}
\usepackage{graphicx}
\usepackage[justification=centering]{caption}
\usepackage{float}
\usepackage{verbatim}
\usepackage{algorithm}
\usepackage{algorithmic}

\begin{document}
	
\title{AST-GIN: Attribute-Augmented Spatial-Temporal Graph Informer Network for Electric Vehicle Charging Station Availability Forecasting}

\author{Ruikang~Luo,~\IEEEmembership{Student Member,~IEEE,}
	    Yaofeng~Song,
	    Liping~Huang,~\IEEEmembership{Member,~IEEE,}
	    Yicheng~Zhang,~\IEEEmembership{Member,~IEEE,}
        and~Rong~Su,~\IEEEmembership{Senior Member,~IEEE}
        
\thanks{Ruikang Luo is affiliated with Continental-NTU Corporate Lab, Nanyang Technological University, 50 Nanyang Avenue, 639798, Singapore Email: ruikang001@e.ntu.edu.sg}
\thanks{Yaofeng Song is affiliated with School of Electrical and Electronic Engineering, Nanyang Technological University, 639798, Singapore Email: song0223@e.ntu.edu.sg@e.ntu.edu.sg}
\thanks{Liping Huang is affiliated with School of Electrical and Electronic Engineering, Nanyang Technological University, 639798, Singapore Email: liping.huang@ntu.edu.sg}
\thanks{Yicheng Zhang is affiliated with Institute for Infocomm Research (I2R), Agency for Science, Technology and Research (ASTAR), 138632, Singapore Email: zhang$\_$yicheng@i2r.a-star.edu.sg}%
\thanks{Rong Su is affiliated with Division of Control and Instrumentation, School of Electrical and Electronic Engineering, Nanyang Technological University, 50 Nanyang Avenue, Singapore 639798. Email: rsu@ntu.edu.sg}
}

\markboth{IEEE TRANSACTIONS ON VEHICULAR TECHNOLOGY}%
{Shell \MakeLowercase{\textit{et al.}}: Bare Demo of IEEEtran.cls for IEEE Journals}

\maketitle

\begin{abstract}
Electric Vehicle (EV) charging demand and charging station availability forecasting is one of the challenges in the intelligent transportation system. With the accurate EV station situation prediction, suitable charging behaviors could be scheduled in advance to relieve range anxiety. Many existing deep learning methods are proposed to address this issue, however, due to the complex road network structure and comprehensive external factors, such as point of interests (POIs) and weather effects, many commonly used algorithms could just extract the historical usage information without considering comprehensive influence of external factors. To enhance the prediction accuracy and interpretability, the Attribute-Augmented Spatial-Temporal Graph Informer (AST-GIN) structure is proposed in this study by combining the Graph Convolutional Network (GCN) layer and the Informer layer to extract both external and internal spatial-temporal dependence of relevant transportation data. And the external factors are modeled as dynamic attributes by the attribute-augmented encoder for training. AST-GIN model is tested on the data collected in Dundee City and experimental results show the effectiveness of our model considering external factors influence over various horizon settings compared with other baselines.

\end{abstract}


\IEEEpeerreviewmaketitle

\section{Introduction}
\IEEEPARstart{A}{ccurate} traffic information forecasting plays an important role in the smart city management. Generally speaking, traffic information contains link speed, traffic flow, vehicle density, travelling time, facility usage condition and so on\cite{luo2020traffic}. As the rapid development of EV technologies, the proportion of EV is growing annually\cite{irle2021global} and Fig.1 shows the world EV sales statistics. However, limited endurance and charging stations, and much longer charging time compared with short refueling time for petrol cars cause serious mileage anxiety for EV drivers\cite{storandt2012cruising}. As one of the most significant infrastructures of EV system, EV charging stations attract more attention recently. Some studies show that EV charging behavior has obvious periodicity\cite{qian2010load}, thereby accurate EV charging station usage condition forecasting system can effectively alleviate range anxiety and improve road efficiency\cite{kondo2013extent}.
\begin{figure}[!htb]
	\centering
	\includegraphics[width=1\linewidth]{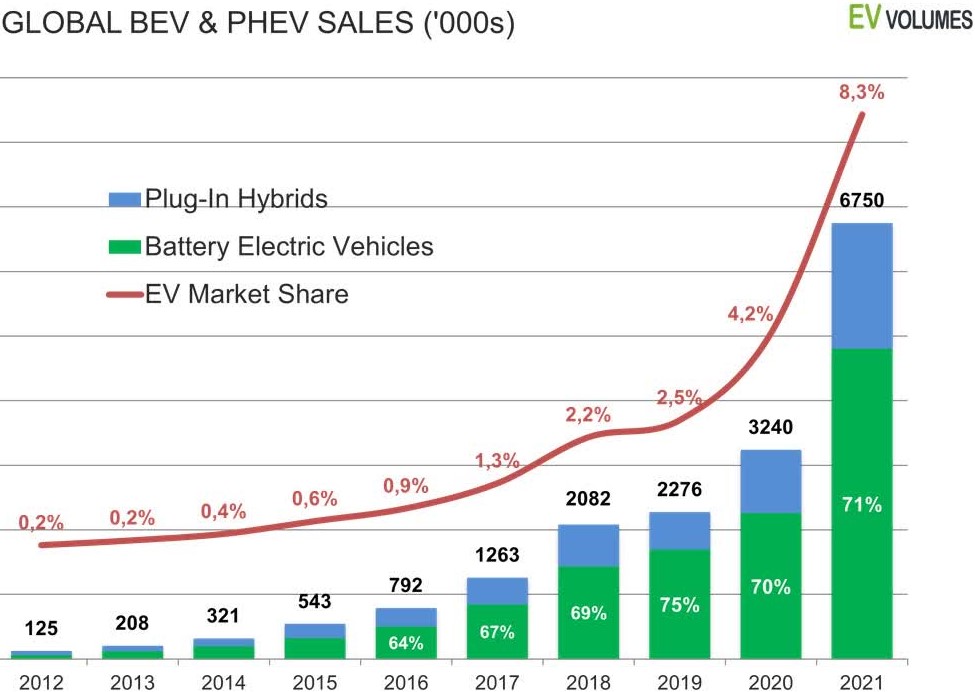}
	\caption{World electric vehicle sales statistics. The global electric vehicle sales increasing 108$\%$ from 4.2$\%$ market share in 2020 to 8.3$\%$ market share in 2021\cite{irle2021global}.}
	\label{fig000}
\end{figure}


Benefit from huge number of smart sensors, real-time station level monitoring has been realized\cite{deilami2011real}. Most canonical facility usage condition prediction methods are dependent on past traffic features to make prediction. However, EV charging station availability is much more complex than other time-series forecasting issue because the future availability not only depends on the historical values, but also influenced by topological relationship and comprehensive external influence\cite{xiong2017optimal}. For example, within the campus or Central Business District (CBD) road section, the usage of charging station will be highly affected by the commute time. An obvious rise of availability can be observed around off-duty time, which is reverse inside residential area, even though two road structures are similar\cite{alface2019electric}. Another example is that bad weather, such as heavy rain, can reduce and delay people's commute, and further affect charging station usage\cite{yan2020ev}. It is quite a challenge to take into consideration of the randomness caused by these external factors\cite{zhu2021ast}.

As the development of deep learning technologies, several forecasting methods has been proposed to solve this issue\cite{huang2017study}, such as Auto-regressive and Integrated Moving Average (ARIMA) method\cite{siami2018comparison}, Convolutional Neural Network (CNN) method, Long Short Term Memory (LSTM) method, GCN method\cite{jiang2021graph} and Transformer based method\cite{yang2020domain}. Each algorithm has its own strengths and limitations. However, most of the models do not have the capability to obtain the augmented attributes during data processing. Correspondingly, perception of external factors is frail. In the second section, detailed introduction of the related work is given.

Comparing with recent related works, we build a novel neural network extracting both spatial-temporal information and external influence to predict the charging station usage condition. The contributions can be concluded as follows:
\begin{itemize}
	\item The novel AST-GIN model containing Attribute Augmentation Unit (A2Unit), GCN Network and Informer Network is proposed to deal with EV charging prediction issue.
	
	\item The proposed AST-GIN model can well consider the dynamic and static external factor influence on the EV charging behavior.
	
	\item To verify the feasibility, AST-GIN and baseline models are tested on one real-world dataset collected at Dundee City. The results show the better prediction capability of AST-GIN over different horizons and metrics.
\end{itemize}

The rest content is arranged as follows: The second section describes related research about deep learning approaches for traffic facility usage forecasting and external factor influence during time-series prediction. The third section illustrates the problem statement and proposed model structure. The fourth section shows detailed experiments with analysis. And the final section summarizes the contribution and possible future plan.

\section{RELATED Research}
In this section, related work on traffic forecasting system and external factor influence on prediction will be introduced.

\subsection{EV charging issue}
Recent research shows that it is a challenge for a fleet of EVs operation, since frequent charging sessions needed\cite{amara2021review}. Alleviate charging stations congestion has become significant for the charging infrastructure management efficiency improvement\cite{bikcora2016prediction}\cite{kim2019predicting}. Two main research directions of EV charging problems are studied recently. One direction focuses on modelling individual EV charging loads and charging stations. The objective is to predict parameters of charging load profiles for smart charging management system\cite{lee2019acn}. Existing studies mainly apply statistical models\cite{flammini2019statistical}, such as Gaussian mixture models\cite{lee2019acn}, and deep learning approaches\cite{majidpour2016forecasting}, such as Hybrid LSTM neural network\cite{hochreiter1997long}\cite{ma2022multistep}, to forecast charging loads at EV charging stations. The other direction analyses modelling and predicting the charging occupancy profile at chargers, which is quite similar with parking availability prediction problem\cite{zhang2020semi}\cite{yang2019deep}. The purpose is to design the scheduling algorithm to allocate EVs among eligible chargers to realize the global or local optimal charging waiting plan\cite{pantelidis2021node}\cite{ma2021optimal}.

\subsection{Canonical forecasting model}
For the traffic forecasting issue, the approaches have undergone several stages and methods could be generally divided into two types: canonical models and deep learning-based models\cite{zhao2020attention}. Canonical forecasting models usually build mathematical models and treat traffic behavior as the conditional process. There are many famous models, such as Historical Average (HA) model, K-nearest Neighbor model, ARIMA model and Support Vector Regression (SVR) model\cite{luo2019traffic}. Most of them consider the trend of data and make the strong assumption that time-series data is stable, which makes them difficult to response the rapid change of inputs\cite{zhao2021domain}.

\subsection{Deep learning forecasting model}
Recently, deep learning-based forecasting methods have been widely applied to solve time-series prediction problem\cite{yang2020real}. Benefit from the capability to extract nonlinear relationships across input sequence, Recurrent Neural Network (RNN) model, Stacked Autoencoding Neural Network (SAE), Gated Recurrent Unit (GRU)\cite{zhang2018combining}, LSTM\cite{shi2015convolutional}, Transformer\cite{vaswani2017attention} and their variants are verified much more efficient to extract the temporal information than canonical forecasting models. To adaptively predict comprehensive traffic conditions, some improving works have been done, such as integrating GCN to extract the spatial dependencies\cite{luo2021deep}\cite{jiang2021graph}.

\begin{figure}[!htb]
	\centering
	\includegraphics[width=1\linewidth]{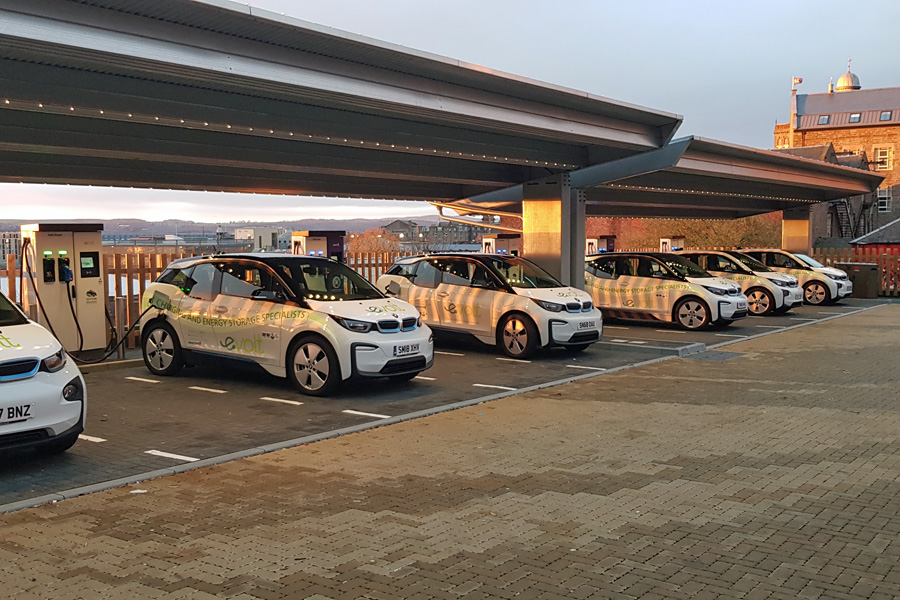}
	\caption{Electric vehicle charging hub opens in Dundee}
	\label{fig08}
\end{figure}

\subsection{External factors in forecasting}
As mentioned above, external factors make influence towards future usage condition of EV stations. To integrate the information of a variety of external information, such as surrounding POIs\cite{wagner2013optimal} and weather conditions\cite{koncar2021probabilistic}, previous studies show us great efforts from multi-source data to specifically designed model structure. In\cite{liao2018deep}, authors proposed the LSTM based structure integrating an encoder to aggregate external information and treat multi-source data as the sequential inputs. In\cite{zhang2018combining}, authors applied the feature fusion technology to process the input weather data for traffic prediction.

In conclusion, existing methods can be further improved by considering external information fluence. Therefore, motivated by related works and the challenge, AST-GIN Network for EV charging station availability forecasting, which integrates both spatial-temporal and external factors as input to enhance model's perception capability during predicting is proposed. In the third section, the architecture and principle of proposed model are illustrated.

\section{METHODOLOGY}
\subsection{Problem Statement}
The purpose of this study is to predict future EV stations usage condition based on historical states and associated information. Based on the priori knowledge introduced, the demand of EVs has a strong periodicity and external factors have a high correlation with the usage of EVs as shown in Fig.3. It is highly potential to find such a mapping function $f$. Thus, without loss of generality, we take the aggregated EV station availability data and two external factors, weather condition and POIs information, as the example to illustrate our proposed model. 
\begin{figure}[!htb]
	\centering
	\includegraphics[width=1\linewidth]{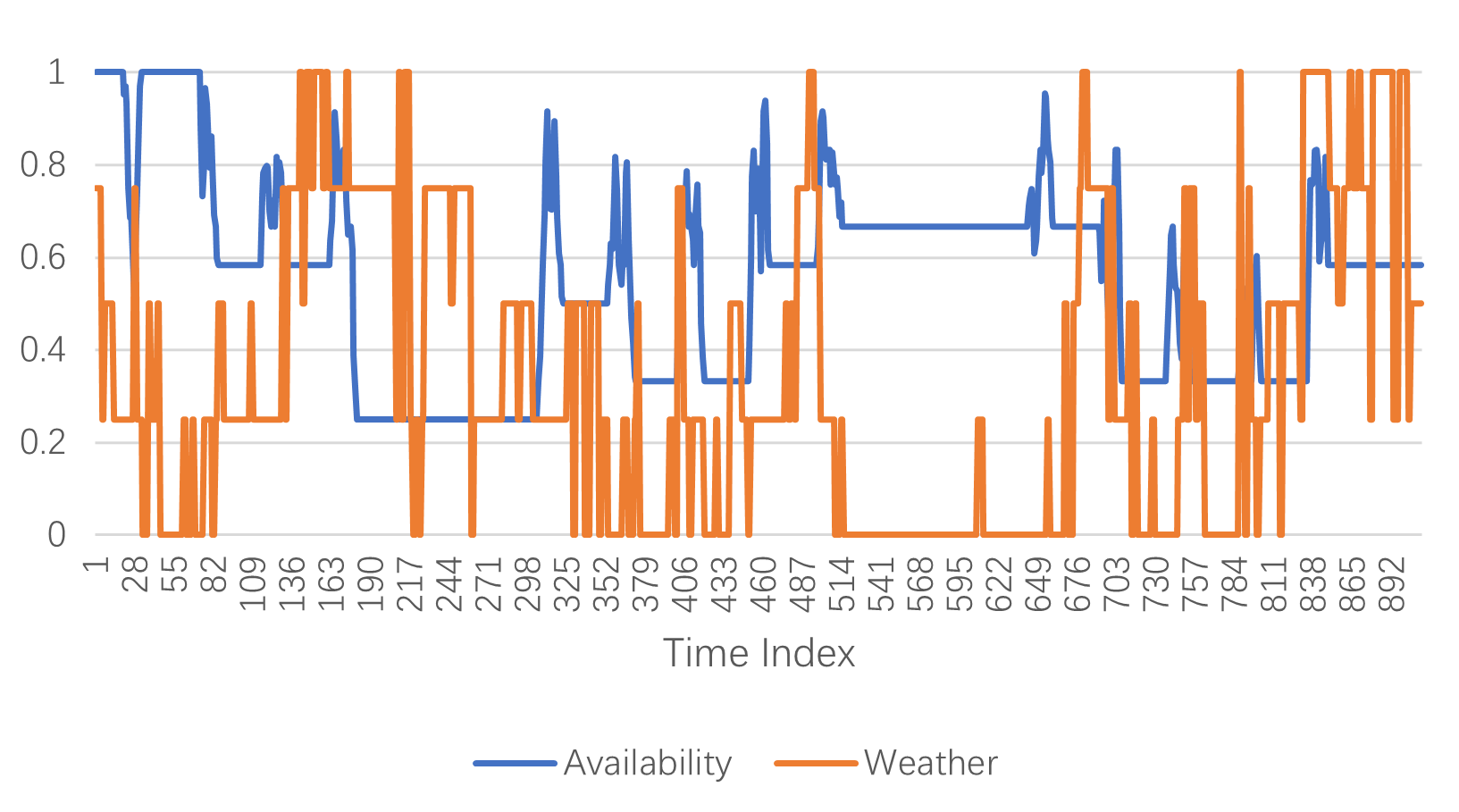}
	\caption{The figure presents the relationship between the weather and the EV charging station availability. The weather is classified into 5 types including sunny, cloudy, foggy, light rainy and heavy rainy which are labeled as 1 to 5. To better present the relationship, the weather data is normalized into the range of 0 to 1. Therefore, the Y-axis for availability data refers to the availability. At the same time, the Y-axis for the weather data refers to different kinds of weather. The higher the value, the worse the weather. The graph shows a general pattern that availability becomes higher when the weather gets better and vice versa. Meanwhile, time index in the figure refers to the recording time.}
	\label{fig18}
\end{figure}

In this paper, the road network is treated as a weighted undirected graph $G=\{V, E\}$. EV stations work as nodes inside the graph, which are notated by $|V| = N$, where $N$ is the number of stations and $E$ represents the graph edge set representing stations connectivity. The corresponding adjacency matrix $A$ $\in$ $R^{N \times N}$ can be constructed based on nodes and edges information. With the road map, based on the latitude and longitude of the charging points, the road distance between EV stations could be estimated. Adjacency matrix elements are calculated using the Gaussian kernel weighting function\cite{zhao2019t}:
\begin{equation}
A_{ab} = 
\left\{
\begin{array}{lr}
exp(-\frac{dist(v_a, v_b)^2}{\sigma^2}),\ dist(v_a, v_b) \leq \kappa &\\
0,\ otherwise &
\end{array}
\right.
\end{equation}
where $dist(v_a, v_b)$ represents distance between station $v_a$ and station $v_b$; $\sigma$ is the standard deviation of $dist(v_a, v_b)$; $\kappa$ is the filter removing small distances.

At time $i$, traffic feature matrix, $X_i$ $\in$ $R^{N \times C}$, contains high-dimension information of EV station availability, where $C$ is the hyperparameter manually defined. Thus, the known $L$ steps historical usage data are defined as $X = \left[ X_{i-L}, X_{i-L+1}, ..., X_i \right]$ and used as partial inputs to predict the next $M$ steps states $\left[\hat{Y}_{i+1}, ..., \hat{Y}_{i+M}\right]$.

Further, the influence of external factors are regarded as affiliated attributes matrix $F$. These factors construct an attribute matrix $\left[ F_{1}, F_{2}, ..., F_{l} \right]$, where $l$ is the dimension of attributes information. At time $i$, the set of $j$-$th$ affiliated information is $F_j = \left[ j_{i-L}, j_{i-L+1}, ..., j_i \right]$.

In conclusion, the issue of EV charging station availability forecasting considering external factors is refined as finding the relationship function $f$ based on the historical usage data $X$, attribute matrix $F$ and road graphic structure $G$, to achieve the future usage values $Y$: 
\begin{equation}
\left[\hat{Y}_{i+1}, ..., \hat{Y}_{i+M}\right] = f (G, X, F)
\label{learning_function}
\end{equation}

In this paper, the traffic parameter studied is the charging station availability. Availability at the $p$-$th$ charging session is aggregated every 30 minutes and could be calculated as:
\begin{equation}
x_{p} = 1-\frac{\sum t_p}{30 \times M_{Connector}}
\label{chargingavailability}
\end{equation}

where $\sum t_p$ is the total charging duration within specific session; $M_{Connector}$ is the charging connector number.

\subsection{AST-GIN Architecture}
In this subsection, the principle of AST-GIN model is introduced in detail. AST-GIN model contains three layers: A2Unit, which can integrate the external information, GCN layer and Informer layer. The historical time-series data and external data are firstly fed into the A2Unit for attribute augmentation. Then, the processed information is fed into GCN layer for spatial information extraction. Finally, Informer layer will take outputs from the GCN layer to extract the temporal dependencies. The overview architecture of AST-GIN is illustrated in Fig.4.
\begin{figure}[!htb]
	\centering
	\includegraphics[width=1\linewidth]{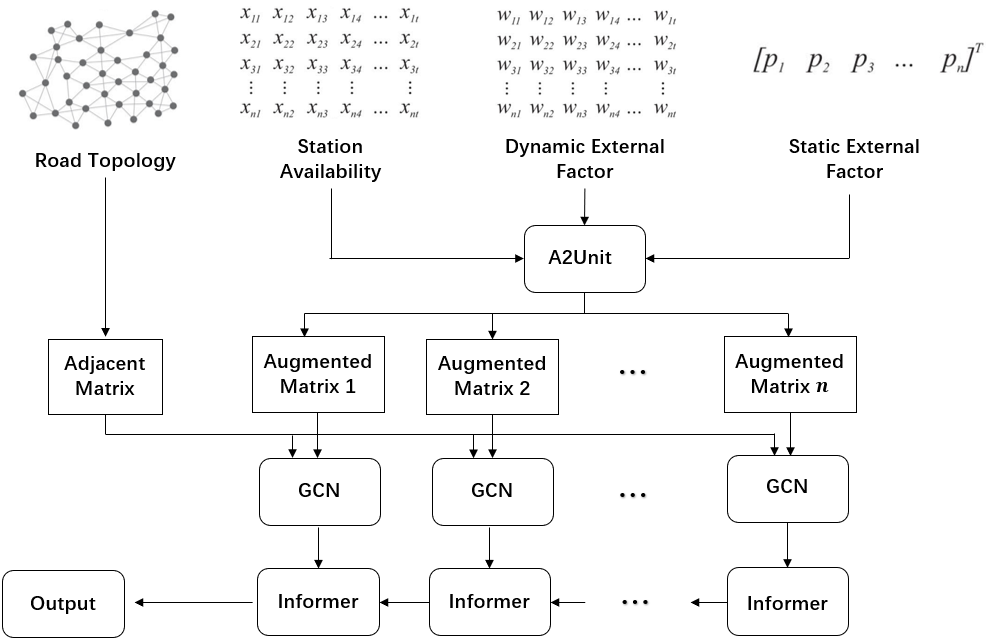}
	\caption{AST-GIN architecture}
	\label{fig19}
\end{figure}

\subsection{Methodology}
\subsubsection{A2Unit}
As mentioned, both historical data and external factors affect EV charging conditions. Thus, different from traditional time-series deep learning forecasting model, additional structure aggregating external factors is needed\cite{liao2018deep}. To comprehensively take external factors influence into consideration, dynamic attribute and static attribute are selected respectively for the objective region. EV stations historical availability tensor $X$, road structure $G$ and two types of attribute matrix are integrated into the A2Unit for augmentation.

We use $\alpha$ $\in$ $R^{N \times p}$ to represent the static attribute matrix containing $p$ categories attributes and $\alpha$ is time-invariant. Similarly $\beta$ $\in$ $R^{N \times (w*t)}$ represents $w$ different dynamic attributes, which are changing over time. To aggregate the cumulative influence of dynamic attributes, $L$ length historical window is selected. Thus, the final augmented matrix processed by A2Unit at time $i$ is stated as:
\begin{equation}
E^i = \left[ X_i, \alpha, \beta^{i-L}, \beta^{i-L+1}, ..., \beta^i \right] 
\end{equation}
where $E^i \in R^{N \times (p+1+w*(L+1))}$, and the same processing procedure is applied for every time stamp inside traffic feature matrix $X$.

\subsubsection{GCN layer}
The distance between vehicle location and the target charging station obviously influence the decision of drivers\cite{quiros2015statistical}. To enhance the understanding of EV charging behavior pattern, spatial dependencies among charging stations are taken into consideration. Some related works have proposed CNN-based neural network to deal with spatial prediction issue\cite{ma2017learning}\cite{ranjan2020city}. However, the distribution of EV charging station connected by the non-Euclidean road network cannot be processed well by CNN. Thus, here, GCN is selected to extract spatial dependencies of input data, which still remains convolutional functionality\cite{kipf2016semi}. The framework of GCN is shown in Fig. 5.

\begin{figure}[!htb]
	\centering
	\includegraphics[width=1\linewidth]{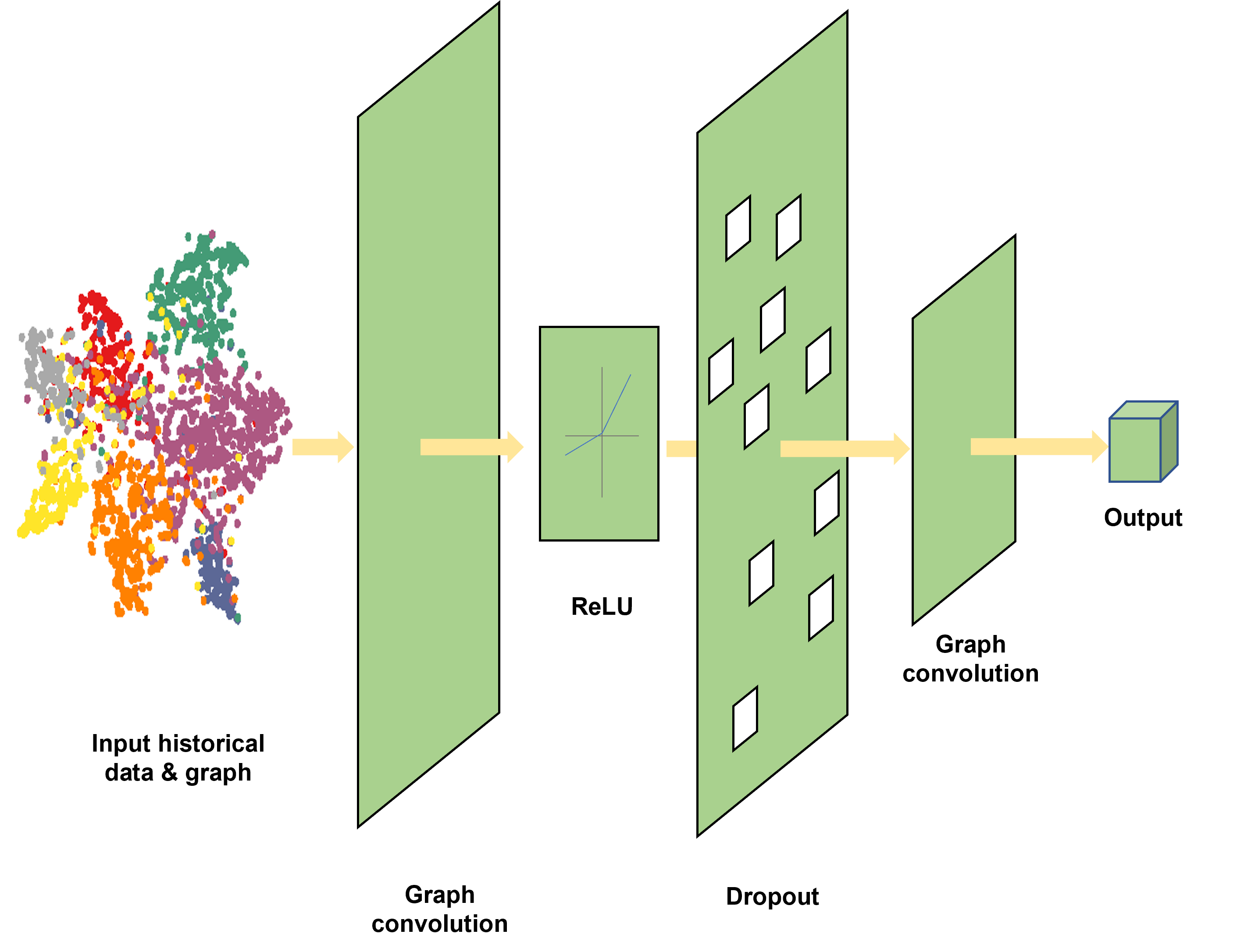}
	\caption{GCN layer architecture}
	\label{fig01}
\end{figure}

In principle, Graph Convolution Neural Network perform the convolution over nodes of the graph to captures spatial information, which is similar with image processing by Convolution Neural Network. 

The general convolution theorem in the spatial domain states that:
\begin{equation}
f*g = \mathcal{F}^{-1}\{\mathcal{F}\{f\}\cdot\mathcal{F}\{g\}\}
\label{convolution}
\end{equation}

where $g$ is the kernel operates on function $f$.

When doing the convolution in the spectral domain, the graph convolution formula can be expressed as:
\begin{equation}
g_{\theta}*x = Ug_{\theta}U^Tx = Ug_{{\theta}'}(\Lambda)U^Tx
\label{graphconvolution}
\end{equation}

where $g_{\theta}$ is the graph convolutional kernel, $U$ is the eigenvector matrix and $\Lambda$ is the diagonal matrix of eigenvalues.

To simplify the computation, people usually make the first order Chebyshev Approximation\cite{kipf2016semi}. And the graph convolution formula now is reorganized as:

\begin{equation}
g_{\theta}*x = \theta(I_n+D^{-\frac{1}{2}}AD^{-\frac{1}{2}})x = \theta(\tilde{D}^{-\frac{1}{2}}\tilde{A}\tilde{D}^{-\frac{1}{2}})x
\label{simplify}
\end{equation}

where $D$ is the diagonal degree matrix, $D_{ii}$ = $\sum_j$ $A_{ij}$; $\tilde{A}$ = A + $I_n$; $\tilde{D}_{ii}$ = $\sum_j$ $\tilde{A}_{ij}$.

Further, GCN convolutional formula could be rewritten as:
\begin{equation}
H^{l+1} = \gamma (\tilde{D}^{-\frac{1}{2}}\tilde{A}\tilde{D}^{-\frac{1}{2}} H^{l}W^{l})
\label{finalconvformula}
\end{equation}

where $\gamma$ is the activation function; $H^{l+1}$ is the $l$-$th$ layer output.

\subsubsection{Informer layer}
It is significant to obtain global temporal dependency while forecasting. As the rapid development of Transformer-like neural network, which employs an encoder-decoder architecture, time-series prediction ability has been improved a lot based on attention mechanism than some canonical deep learning methods\cite{tsai2019transformer}, such as GRU and LSTM. Thus, one latest variant of Transformer, Informer\cite{zhou2021informer}, is applied here as the temporal extraction layer to understand the global sequence. The structure of the Informer is shown as Fig.6.

\begin{figure}[!htb]
	\centering
	\includegraphics[width=1\linewidth]{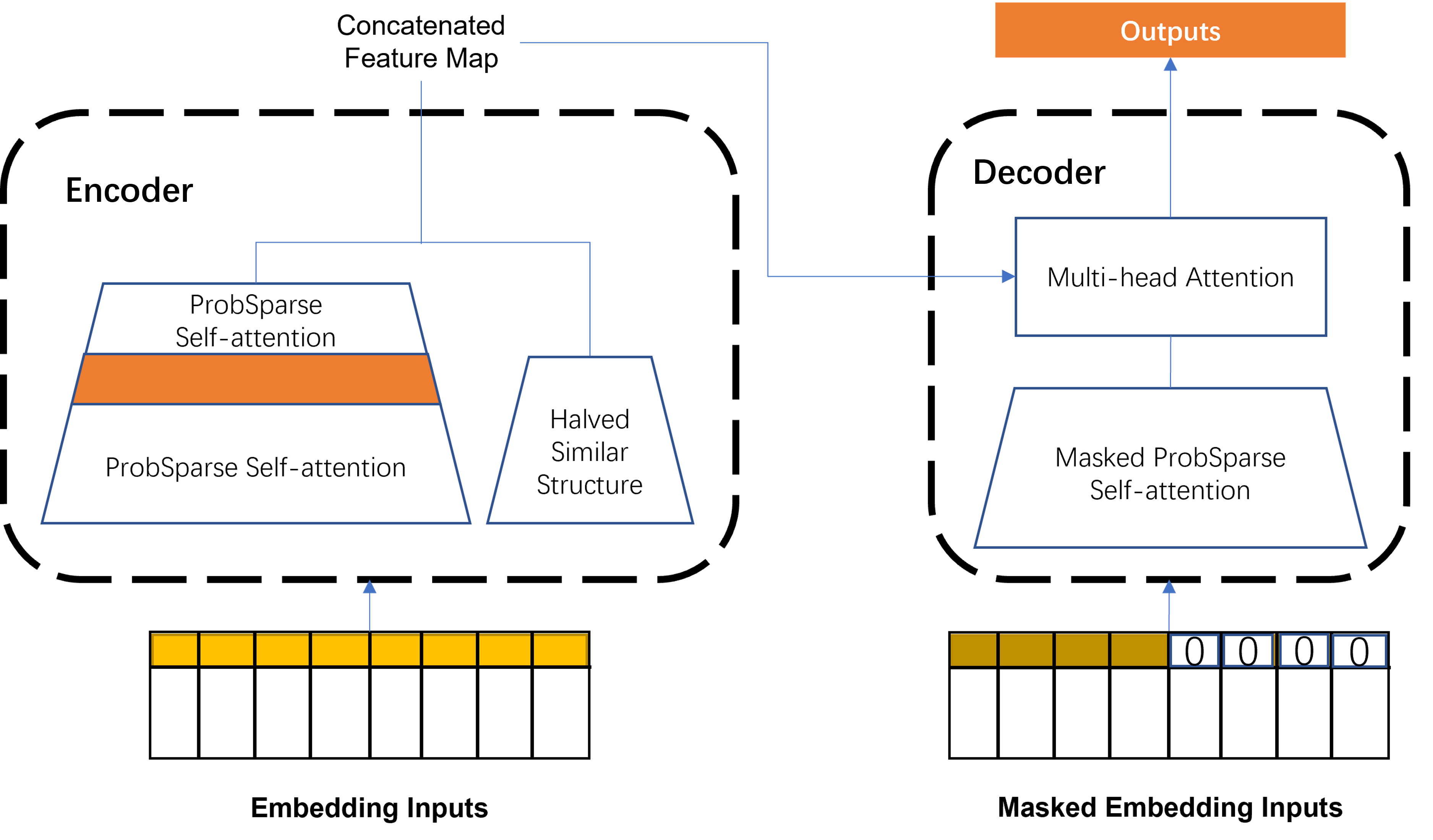}
	\caption{Temporal Informer layer structure}
	\label{fig02}
\end{figure}

\subsection{Loss Function}
During model training, the objective is to eliminate the gap between the ground truth and predicted value. Thus, the loss function can be written as:
\begin{equation}
loss = \|Y_t - \hat{Y_t}\| + \lambda L_{reg}
\label{lossfunction}
\end{equation}

The $Y_t$ and $\hat{Y_t}$ represent the recorded value and the predicted value respectively. $\lambda L_{reg}$\cite{bilgic2014fast} is the L2 regularization term to avoid overfitting during training.

\section{EMPIRICAL ANALYSIS}
To evaluate AST-GIN model performance, necessary experiments have been done on the EV charging station availability dataset. We choose 5 efficient time-series forecasting baseline models for comparison. During the experiment, the performance of AST-GIN model with static external factor only, the model with dynamic external factor only and the model with both static and dynamic factors would be evaluated separately.

\subsection{Data Collection}
Dundee EV charging dataset: this dataset (https://data.dundeecity.gov.uk/dataset/ev-charging-data) is a record of the EV charging behaviors in Dundee, Scotland. There are 57 charging points in Dundee which could be divided into 3 types including slow chargers, fast chargers and rapid chargers. Totally 3 valid datasets recorded in 3 different time periods including 01/09/17 to 01/12/17, 02/12/17 to 02/03/18 and 05/03/18 to 05/06/18 are accessible. Meanwhile, the geographical locations of all the charging points are also provided and they are shown in Fig.7.

\begin{figure}[!htb]
	\centering
	\includegraphics[width=1\linewidth]{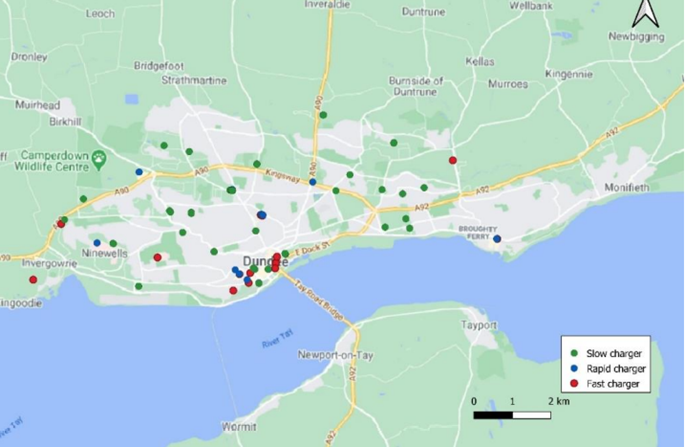}
	\caption{\textbf{Locations of EV charging points.} All the 57 charging points are included in the figure in which green points refer to slow chargers, blue points refer to rapid chargers and red point refer to fast chargers.}
	\label{fig06}
\end{figure}

In the present study, the dataset we used is recorded during 05/03/18 to 05/06/18. There are 16773 charging sessions recorded in total and each of the charging session record contains charging point ID, charging connector ID, starting and ending charging time, total consumed electric power, geographical location, and the type of the charging point. There are 40 slow chargers with 5894 charging sessions recorded, 8 fast chargers with 1416 charging sessions recorded and 9 rapid chargers with 9463 charging sessions recorded.

Moreover, Dundee weather data is available, which is a weather record in Dundee city. The weather is recorded every hour and each record includes the general description of the weather, temperature, wind, humidity and barometer.

\subsection{Data Preprocessing}
\subsubsection{Static External Factors}
We classify the surroundings of the charging points into 8 types including transportation services, catering services, shopping services, education services, accommodation, medical services, living services and other. The category of surroundings with the largest proportion would be labeled as the POI value of a charging point based on the geographical location of it.

\subsubsection{Dynamic External Factors}
The weather of Dundee is divided into 5 types which includes sunny, cloudy, foggy, light rainy and heavy rainy with different labels from 1 to 5. Since the time interval of the weather data from the source dataset is 1 hour, the weather in the covered period is regarded as the same which means that if the weather at 17:50 is recorded as sunny, the weather at 18:20 would be labeled as sunny.

\subsection{Hyperparameters}
In this study, a 3-layer GCN structure is used. For each Informer block, the number of encoder layer is 2 while the number of decoder layer is 3. During the training, 50\% data is randomly divided for training, 33\% data is selected for evaluation and the rest 17\% data is utilized for testing purpose. The network is optimized using Adam optimizer. The learning rate starts from  $e^{-4}$ and decays 10 times smaller every 2 epochs. The total number of epochs is 50 with an early stop determination. The batch size is chosen as 32 and the whole network is trained on a GPU RTX3060.

\subsection{Evaluation Metrics}
In this study, 5 commonly used metrics are selected to evaluate models forecasting performance.

(1) Root Mean Squared Error (RMSE):
\begin{equation}
RMSE = \sqrt{\frac{1}{MN}\sum_{j=1}^M\sum_{i=1}^{N}(y_i^j-\hat{y_i^j})^2}
\label{rmse}
\end{equation}

(2) Coefficient of Determination ($R^2$):
\begin{equation}
R^2=1-\frac{\sum_{j=1}^M\sum_{i=1}^N{(y_i^j-\hat{y_i^j})}^2}{\sum_{j=1}^M\sum_{i=1}^N{(y_i^j-\bar{Y})}^2}
\label{R2}
\end{equation}

(3) Explained Variance Score (var):
\begin{equation}
var=1-\frac{Var\{Y-\hat{Y}\}}{Var\{Y\}}
\label{var}
\end{equation}

(4) Mean Absolute Error (MAE):
\begin{equation}
MAE = \frac{1}{MN}\sum_{j=1}^M\sum_{i=1}^{N}| y_i^j-\hat{y}_i^j |
\label{mae}
\end{equation}

(5) Accuracy:
\begin{equation}
Accuracy = 1-\frac{{\|Y-\hat{Y}\|}_F}{{\|Y\|}_F}
\label{accuracy}
\end{equation}

where $y_i^j$ and $\hat{y}_i^j$ represent the ground truth availability and predicted one for the $i-th$ charging station at time $j$. $M$ is time instant number; $N$ is charging station number; $Y$ and $\hat{Y}$ are the set of $y_i^j$ and $\hat{y}_i^j$ respectively; $\bar{Y}$ is the average of $Y$.

\subsection{Baseline Settings}
As far as we know, there is no directly published models for this specific problem, hence typical models are selected for comparison. Compare the proposed model with the typical sequence model of GRU, LSTM, Transformer and Informer involved in our model, we can testify the importance of incorporating the spatial dependencies captured by GCN. Expecitally, when we compare the proposed model with the Informer, the significance of GCN in the architecture can be verified. The baselines are described as follows:

\begin{itemize}	
	\item GRU: The commonly used time-series model, which has been proven effective in traffic prediction problem and can alleviate the problem of gradient explosion and vanishing.

    \item LSTM: Together with GRU, they are two popular variants of RNN. LSTM has a more complex structure than GRU.
	
	\item Transformer: The classic Transformer model with the self-attention mechanism\cite{vaswani2017attention}.
	
	\item Informer: A new Transformer variant proposed to process long sequence prediction issue without spatial dependencies extraction.	
	
	\item STTN: A new proposed framework utilizes two transformer blocks to capture both spatial and long-range bidirectional temporal dependencies across multiple time steps\cite{xu2020spatial}.

\end{itemize}

\subsection{Experiment Results}
We use 5 state of art baselines to compare the performance with our AST-GIN model, including GRU, LSTM, Transformer, Informer and STTN. Based on the 30-minute time interval of the EV charging availability dataset, we deploy selected models to predict the availability in the next 30-min, 60-min, 90-min and 120-min horizons. The numerical results are shown in TABLE I.

\begin{table*}[!htbp]
	\centering
	\caption{Forecasting results of AST-GIN and baseline models}
	\resizebox{0.95\textwidth}{!}{%
		\begin{tabular}{|c|c|c|c|c|c|c|c|c|c|}
			\hline
			
			\multirow{2}*{Horizon(minutes)} & \multirow{2}*{Metric} &\multirow{2}*{GRU} &\multirow{2}*{LSTM} &\multirow{2}*{Transformer} &\multirow{2}*{Informer} &\multirow{2}*{STTN} &\multicolumn{3}{|c|}{AST-GIN}\\
			\cline{8-10}
			{}&{}&{}&{}&{}&{}&{}&{POI}&{Weather}&{POI+Weather}\\
			\hline
			
			\multirow{5}*{30}&{RMSE}&0.1616&0.2325&0.2803&0.2062&0.171&0.1219&0.1224&\textbf{0.1174}\\
			\cline{2-10}
			&{$R^2$}&0.7675&0.4988&0.4001&0.6623&0.7183&0.8778&0.8709&\textbf{0.8803}\\
			\cline{2-10}
			&{var}&0.7597&0.4897&0.3764&0.65&0.7175&0.8778&0.8704&\textbf{0.8801}\\
			\cline{2-10}
			&{MAE}&0.1017&0.1692&0.2338&0.1565&0.133&0.0748&0.0759&\textbf{0.067}\\
			\cline{2-10}
			&{Accuracy}&0.7513&0.6439&0.5598&0.7239&0.752&0.8328&0.8322&\textbf{0.8388}\\
			\hline

			\multirow{5}*{60}&{RMSE}&0.1819&0.2312&0.2826&0.2362&0.222&0.1464&0.1471&\textbf{0.1438}\\
			\cline{2-10}
			&{$R^2$}&0.684&0.5074&0.3935&0.5477&0.6284&0.8194&0.8147&\textbf{0.8227}\\
			\cline{2-10}
			&{var}&0.6798&0.494&0.3718&0.5367&0.6266&0.8194&0.8147&\textbf{0.8225}\\
			\cline{2-10}
			&{MAE}&0.1186&0.1753&0.2358&0.1807&0.1699&0.0872&0.0864&\textbf{0.0757}\\
			\cline{2-10}
			&{Accuracy}&0.7183&0.6442&0.5543&0.6789&0.6782&0.8004&0.7994&\textbf{0.8037}\\
			\hline

			\multirow{5}*{90}&{RMSE}&0.2269&0.2336&0.2848&0.2613&0.2118&0.1682&\textbf{0.1674}&0.1687\\
			\cline{2-10}
			&{$R^2$}&0.5362&0.496&0.3335&0.4806&0.5718&0.7652&\textbf{0.7653}&0.7605\\
			\cline{2-10}
			&{var}&0.5085&0.485&0.3662&0.4695&0.5634&0.7641&\textbf{0.7652}&0.7604\\
			\cline{2-10}
			&{MAE}&0.1548&0.1741&0.2377&0.1976&0.1683&\textbf{0.0957}&0.0982&0.1017\\
			\cline{2-10}
			&{Accuracy}&0.6508&0.6406&0.5491&0.6581&0.693&0.7713&\textbf{0.7731}&0.7713\\
			\hline		
			
			\multirow{5}*{120}&{RMSE}&0.2327&0.2345&0.2869&0.2828&0.3246&\textbf{0.1844}&0.1852&0.1851\\
			\cline{2-10}
			&{$R^2$}&0.5141&0.4734&0.3273&0.4535&0.5518&\textbf{0.7153}&0.7138&0.7134\\
			\cline{2-10}
			&{var}&0.4809&0.4657&0.3622&0.3922&0.5518&\textbf{0.7153}&0.7131&0.7131\\
			\cline{2-10}
			&{MAE}&0.1556&0.1796&0.2396&0.2118&0.164&0.1125&\textbf{0.1106}&0.1123\\
			\cline{2-10}
			&{Accuracy}&0.6418&0.6392&0.5437&0.6293&0.6889&\textbf{0.7507}&0.7496&0.7496\\
			\hline
			
		\end{tabular}%
	}
	\label{table1}
\end{table*}

Fig.8. to Fig.16. show the visualization results for the AST-GIN model under static, dynamic and combined external factors over different prediction horizons. The blue lines represent the ground truth collected in Dundee and the orange lines represent the predicted values. It can be observed that the overall forecasting capability is stable over time. The combination of both static and dynamic factors leads higher prediction accuracy in the relative short-term. And the difference among static factors only, dynamic factors only and the combination becomes negligible in the relative long-term.
\begin{figure}[!htb]
	\centering
	\includegraphics[width=1\linewidth]{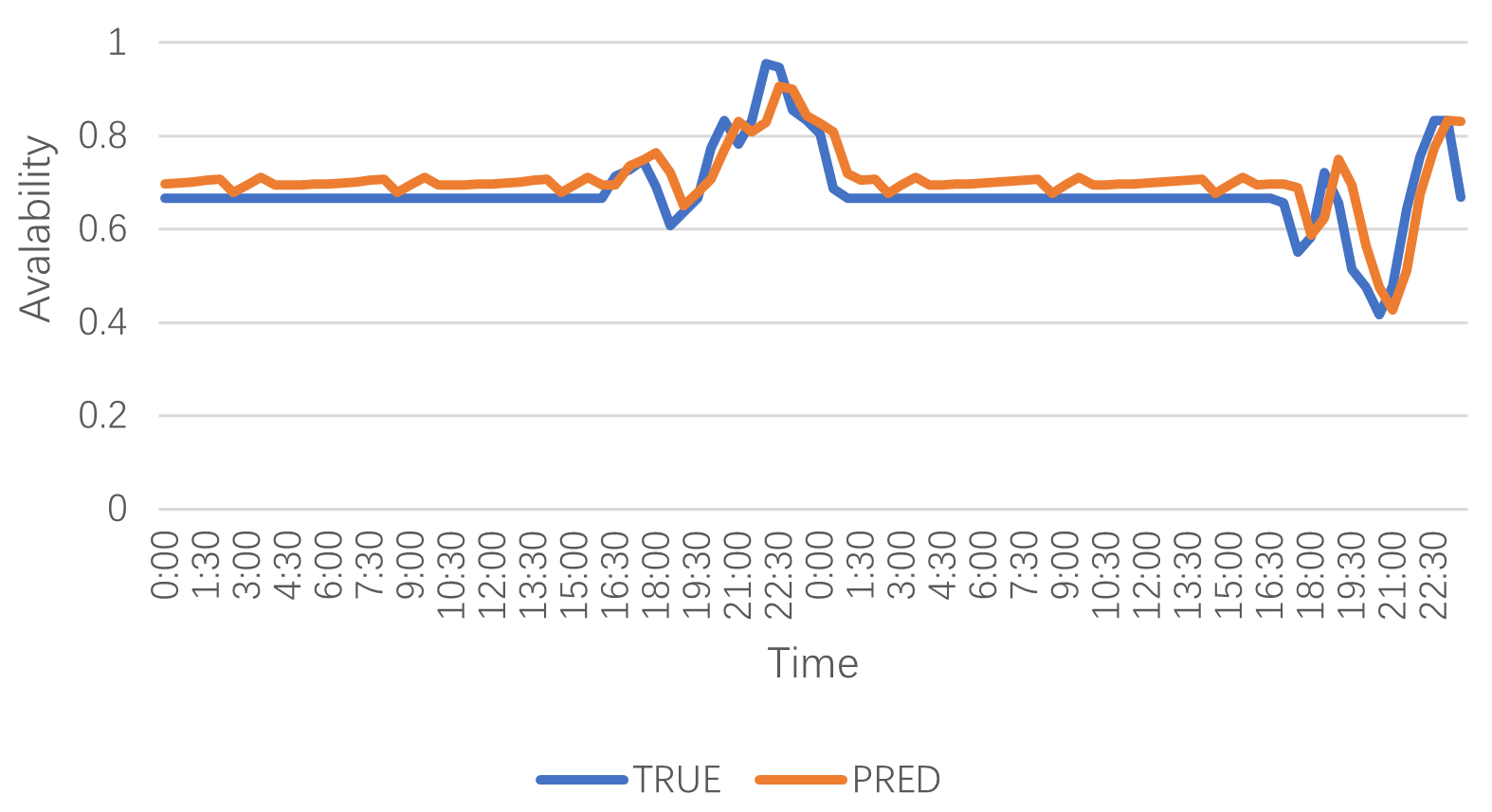}
	\caption{Visualization results for static factor of 30-min horizon}
	\label{fig33}
\end{figure}

\begin{figure}[!htb]
    \centering
    \includegraphics[width=1\linewidth]{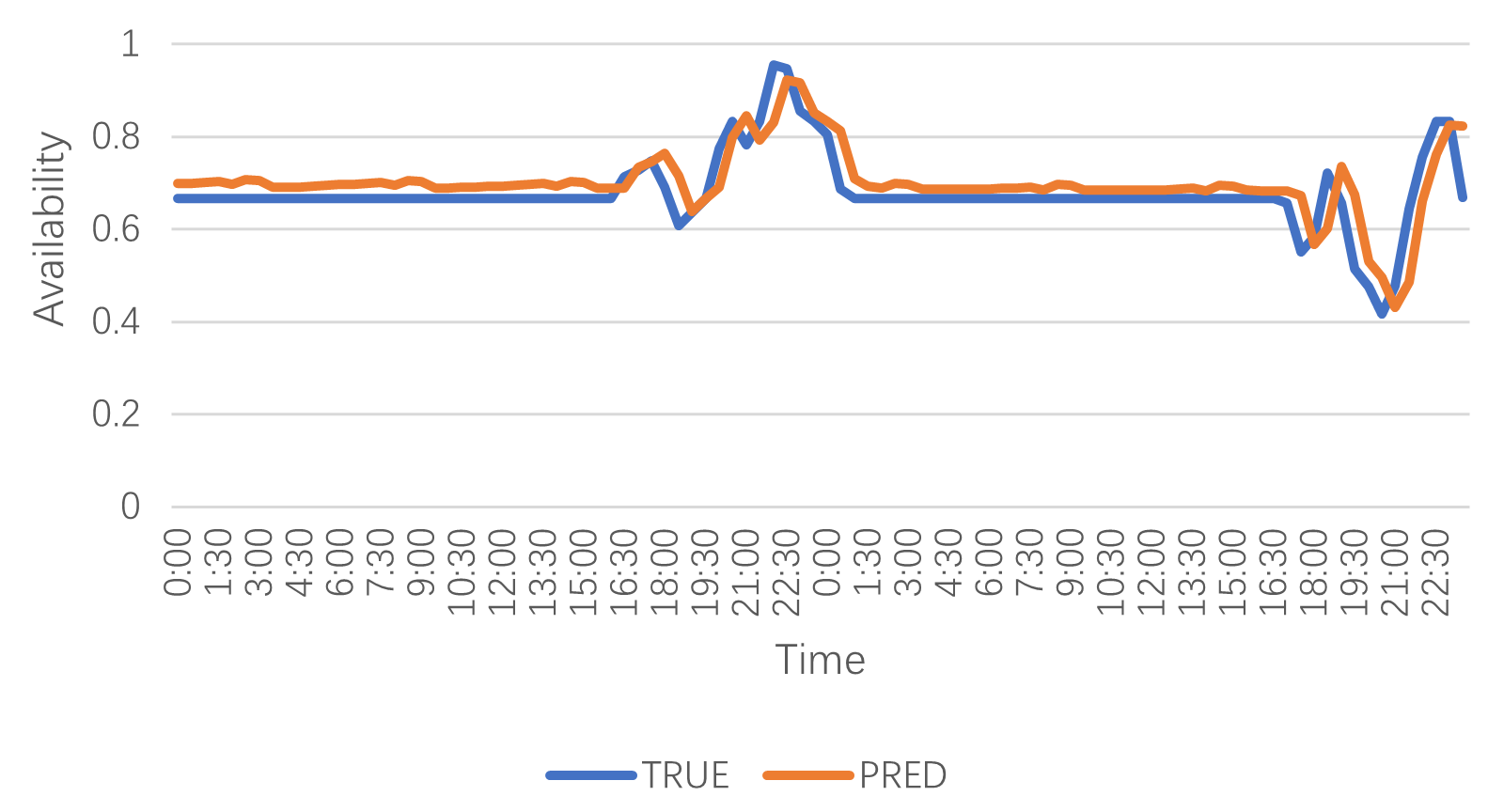}
    \caption{Visualization results for static factor of 60-min horizon}
    \label{fig34}
\end{figure}

\begin{figure}[!htb]
	\centering
	\includegraphics[width=1\linewidth]{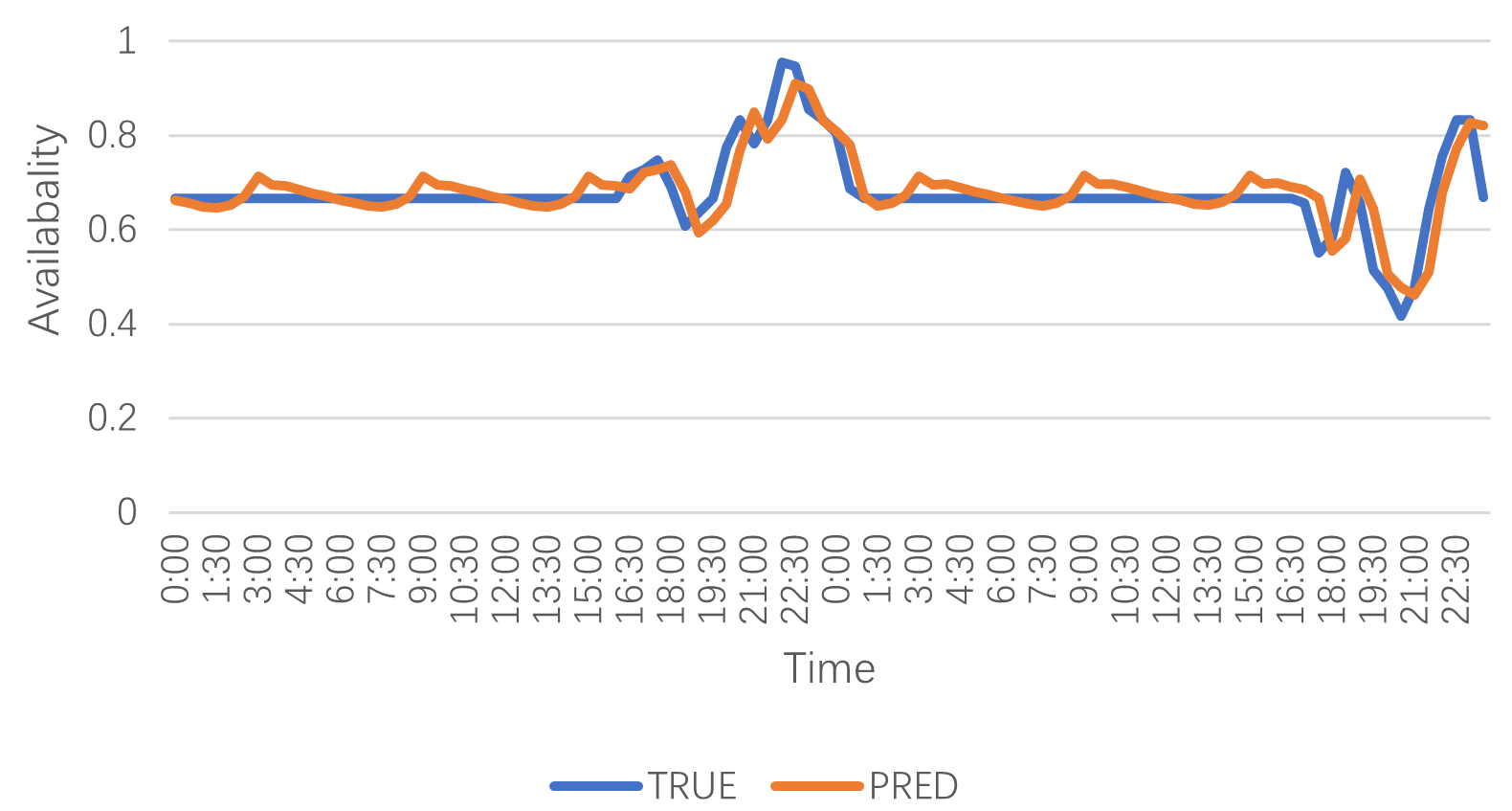}
	\caption{Visualization results for static factor of 90-min horizon}
	\label{fig35}
\end{figure}

\begin{figure}[!htb]
	\centering
	\includegraphics[width=1\linewidth]{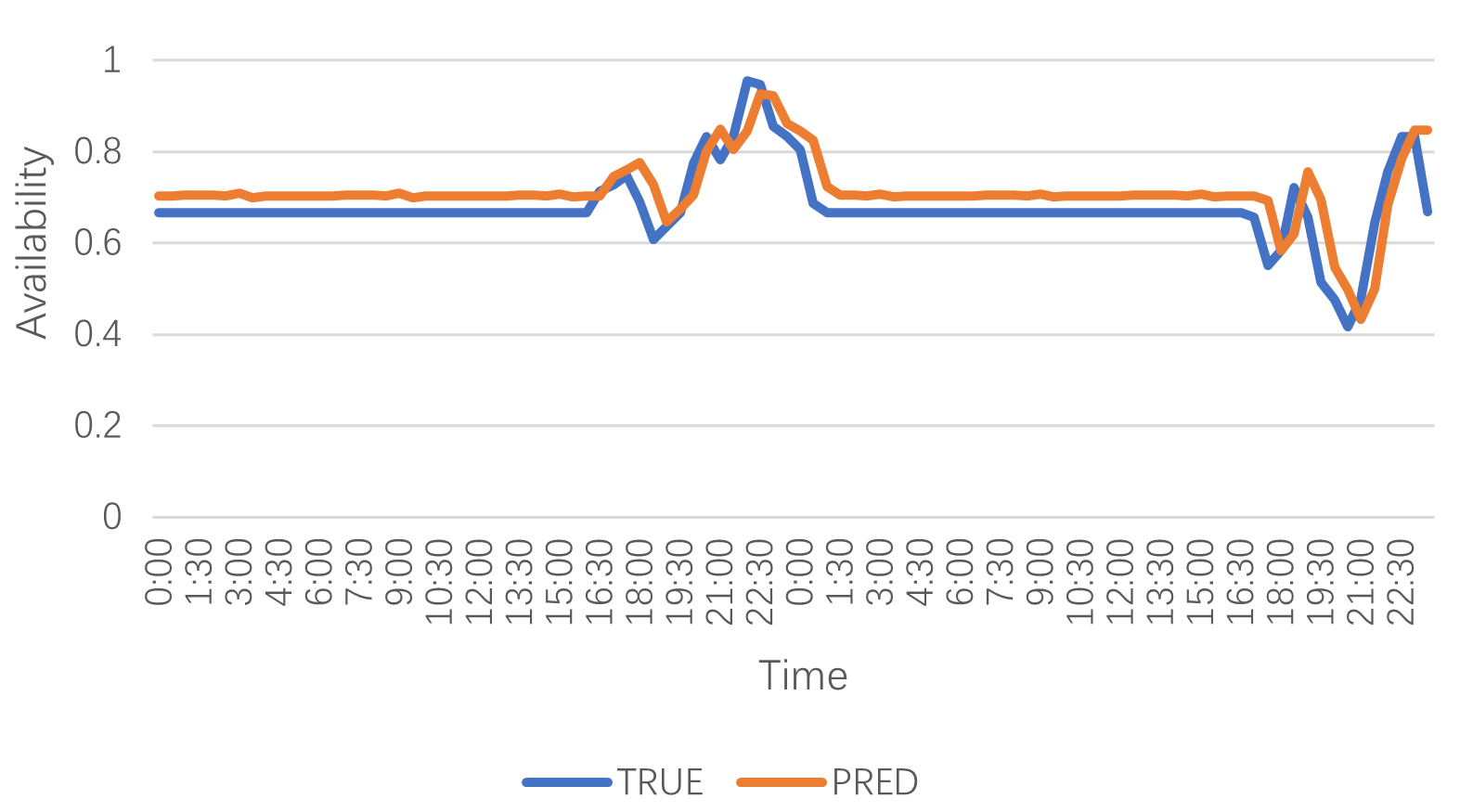}
	\caption{Visualization results for dynamic factor of 30-min horizon}
	\label{fig36}
\end{figure}

\begin{figure}[!htb]
	\centering
	\includegraphics[width=1\linewidth]{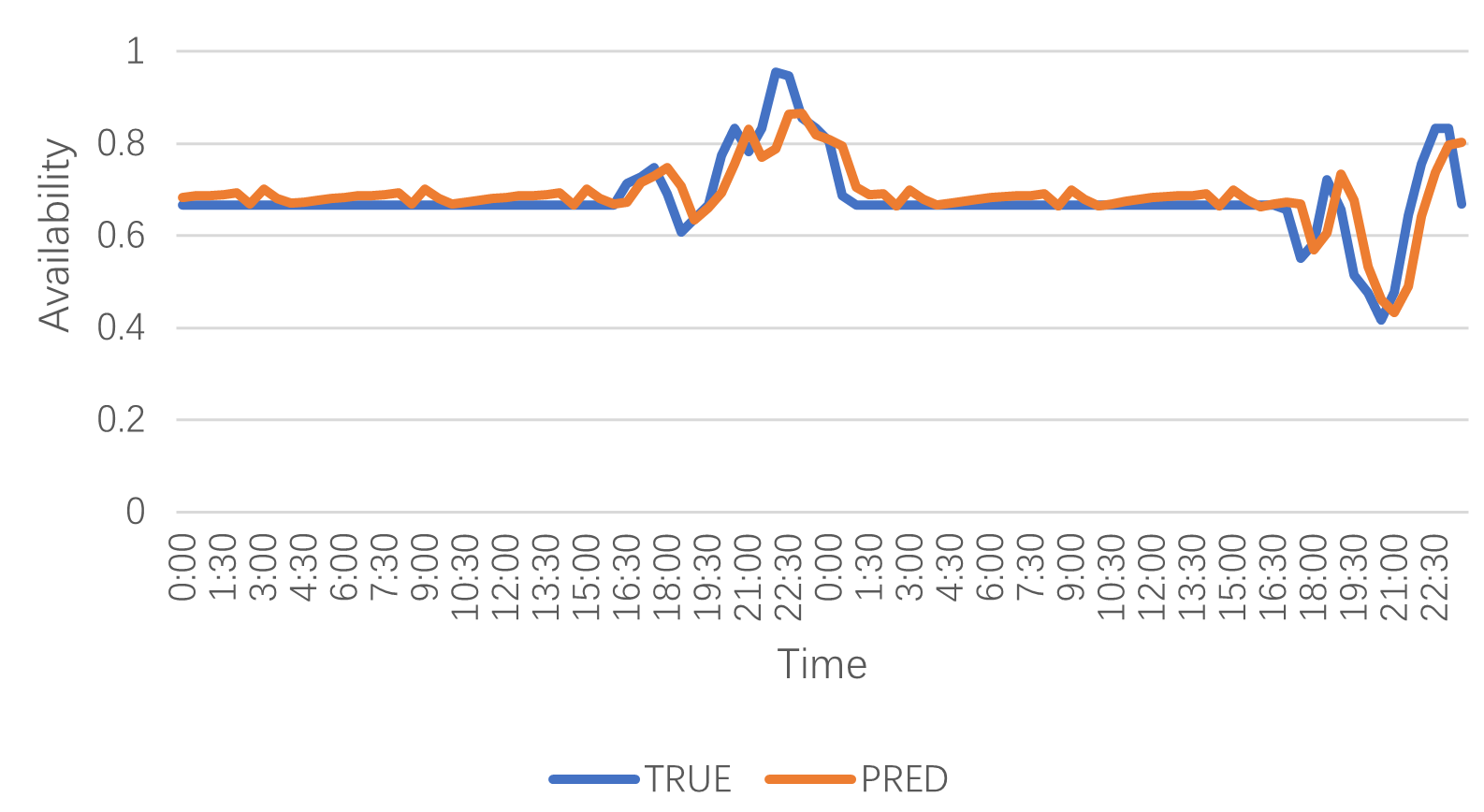}
	\caption{Visualization results for dynamic factor of 60-min horizon}
	\label{fig37}
\end{figure}

\begin{figure}[!htb]
	\centering
	\includegraphics[width=1\linewidth]{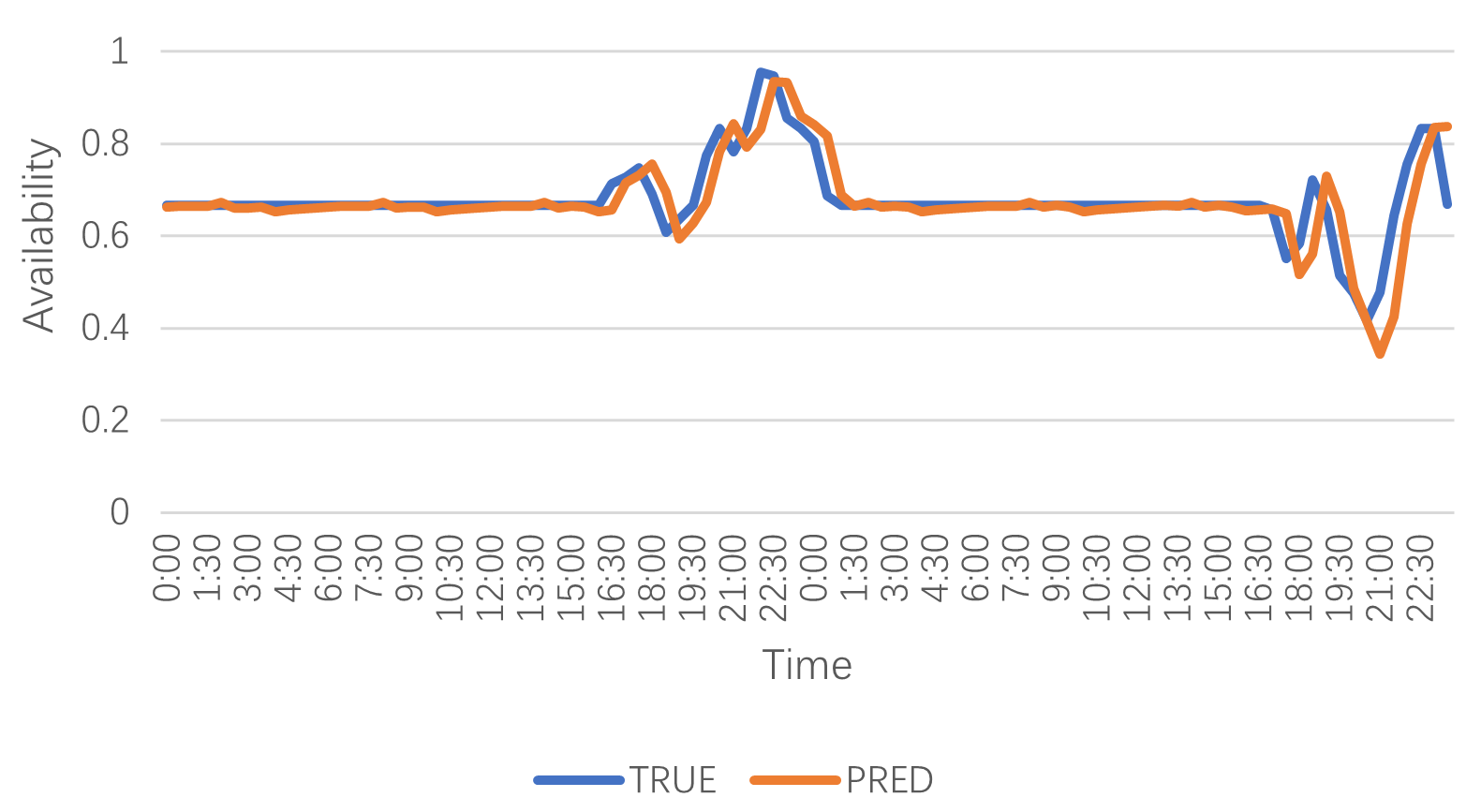}
	\caption{Visualization results for dynamic factor of 90-min horizon}
	\label{fig38}
\end{figure}

\begin{figure}[!htb]
	\centering
	\includegraphics[width=1\linewidth]{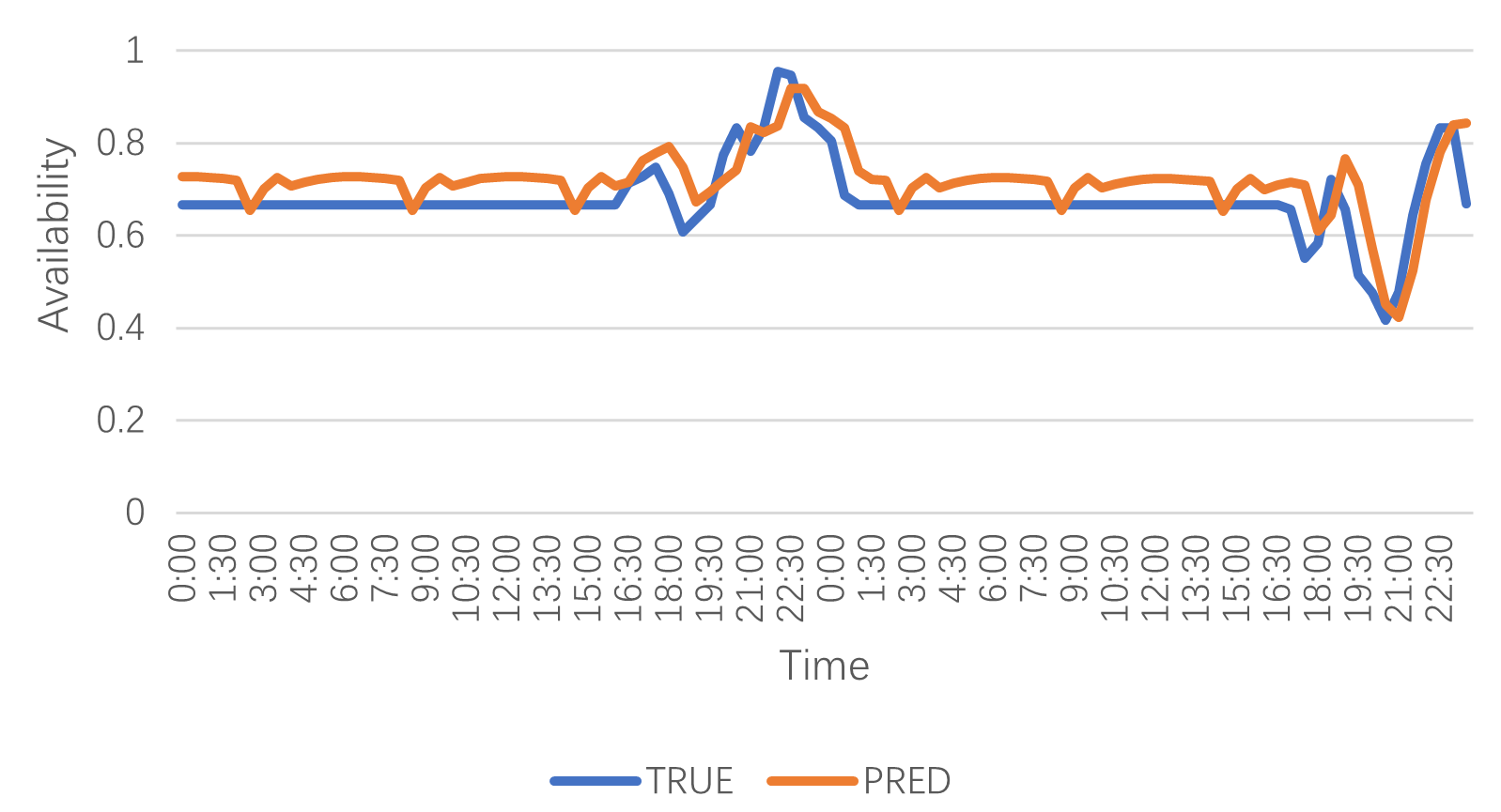}
	\caption{Visualization results for combined factor of 30-min horizon}
	\label{fig39}
\end{figure}

\begin{figure}[!htb]
	\centering
	\includegraphics[width=1\linewidth]{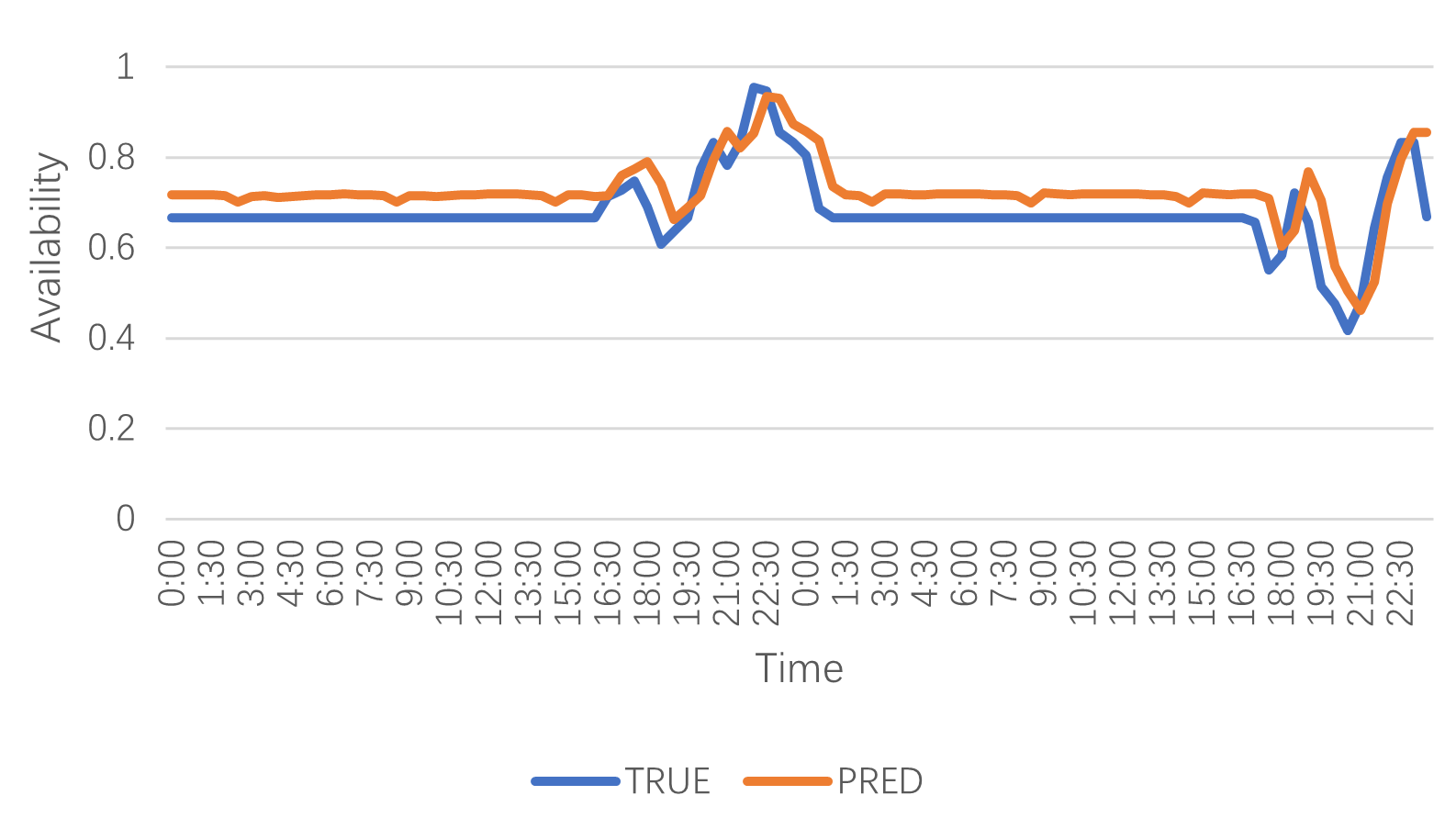}
	\caption{Visualization results for combined factor of 60-min horizon}
	\label{fig40}
\end{figure}

\begin{figure}[!htb]
	\centering
	\includegraphics[width=1\linewidth]{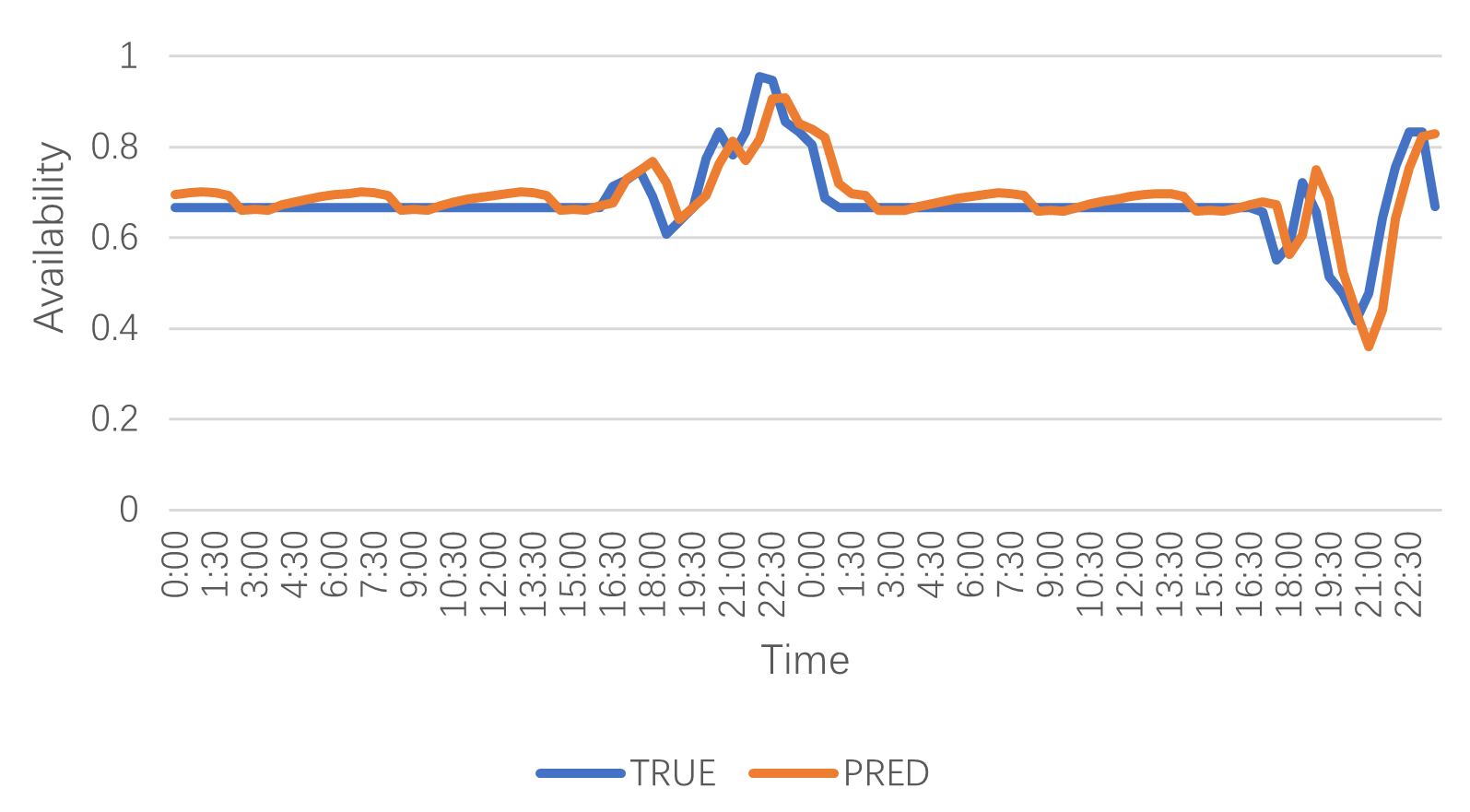}
	\caption{Visualization results for combined factor of 90-min horizon}
	\label{fig41}
\end{figure}

In the short-term EV charging station availability forecasting, horizons of 30 minutes for example, AST-GIN model effectively captures the temporal dependance of the data and outperform other baseline models with a higher accuracy.  In the 90 minutes forecasting, AST-GIN model with the dynamic external factor achieves an accuracy of 0.8388, while the best model, STTN, among the baseline models achieve an accuracy of 0.7520. AST-GIN outperforms STTN in the 90 minutes horizon with a 11.54\% higher accuracy. In the long-term forecasting, 120 minutes for example, AST-GIN still has the best performance with the highest accuracy compared to the other baseline models. The accuracy of AST-GIN, 0.7507, is 8.97\% higher than the best model, STTN, among baseline models. The consolidated result is shown in Fig.17.

Among the 3 kinds of used external factors which includes static factors only, dynamic factors only and the combination of both of the factors, the use of external factors combination leads to a better performance in general.

\begin{figure}[!htb]
	\centering
	\includegraphics[width=1\linewidth]{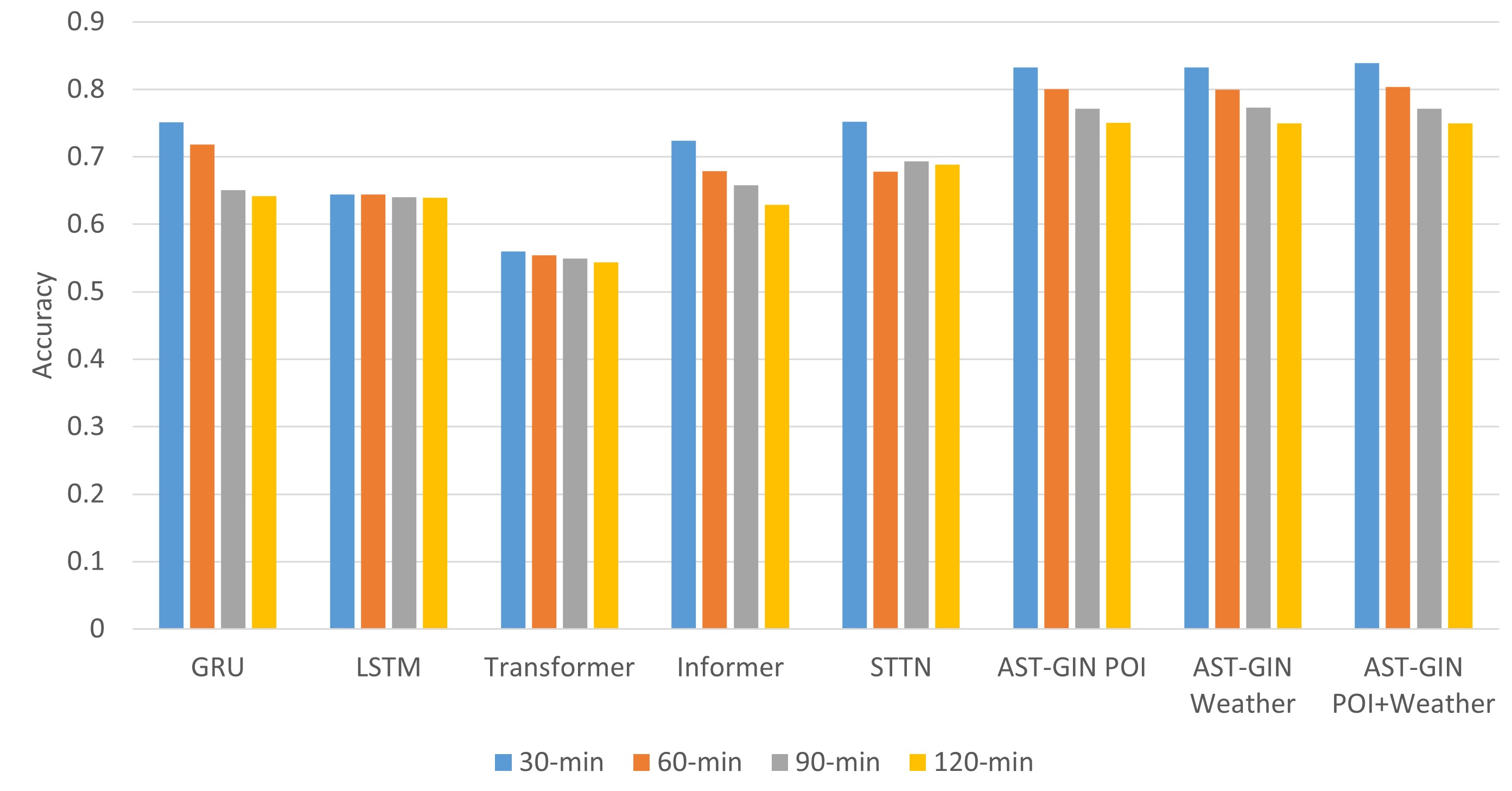}
	\caption{Accuracy Statistics of All Models}
	\label{fig03}
\end{figure}

\subsection{Perturbation Analysis and Robustness}
Inevitably noise during the data collection, such as sensor error and system delay, may undermine the model forecasting capability. To verify the noise immunity of AST-GIN model, the model robustness is tested via perturbation experiments. The normalized random noise, which obey the Gaussian distribution, is added to the data to check the robustness. The resultant fluctuation is small.

\section{CONCLUSIONS}
In this paper, a deep learning model AST-GIN is proposed and verified for EV charging station availability forecasting considering external factors influence. The model contains A2Unit layer, GCN layer and Informer layer to augment time-series traffic features and extract spatial-temporal dependencies of EV charging station usage condition. AST-GIN and baselines are tested on the data collected in Dundee City. Experiments show that AST-GIN has the better forecasting capability over various horizons and metrics. To summarize, AST-GIN can effectively consider comprehensive external attribute influence and predict EV charging station usage condition. Future plan regarding robustness is ongoing to further improve the predictive system.

\section*{ACKNOWLEDGMENT}

This study is supported under the RIE2020 Industry Alignment Fund – Industry Collaboration Projects (IAF-ICP) Funding Initiative, as well as cash and in-kind contribution from the industry partner(s).

\ifCLASSOPTIONcaptionsoff
  \newpage
\fi


\singlespacing
\bibliographystyle{IEEEtran}
\bibliography{autosam}


\begin{IEEEbiography}[{\includegraphics[width=1in,height=1.25in,clip,keepaspectratio]{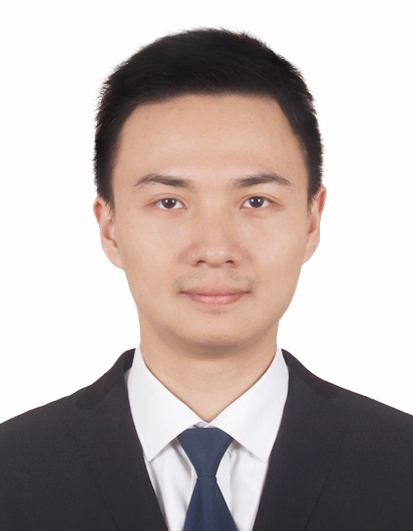}}]{Ruikang Luo}
    received the B.E. degree from the School of Electrical and Electronic Engineering, Nanyang Technological University, Singapore. He is currently currently pursuing the Ph.D. degree with the School of Electrical and Electronic Engineering, Nanyang Technological University, Singapore. His research interests include long-term traffic forecasting based on spatiotemporal data and artificial intelligence.
\end{IEEEbiography}

\begin{IEEEbiography}[{\includegraphics[width=1in,height=1.25in,clip,keepaspectratio]{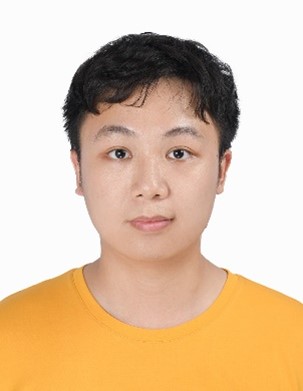}}]{Yaofeng Song}
	received the bachelor degree from the school of Automation Science and Engineering in South China University of Technology. Currently he is a Msc student in the school of Electrical and Electronic Engineering in Nanyang Technological University, Singapore. His research interests invlove deep learning based traffic forecasting.
\end{IEEEbiography}

\begin{IEEEbiography}[{\includegraphics[width=1in,height=1.25in,clip,keepaspectratio]{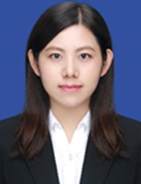}}]{Liping Huang}
	Huang Liping obtained her Ph. D, and Master of Computer Science from Jilin University in 2018 and 2014, respectively. She has been working as a research fellow at Nanyang Technological University since 2019 June. Dr. Huang’s research interests include spatial and temporal data mining, mobility data pattern recognition, time series prediction, machine learning, and job shop scheduling. In the aforementioned areas, she has more than twenty publications and serves as the reviewer of multiple journals, such as IEEE T-Big Data, IEEE T-ETCI, et al.
\end{IEEEbiography}

\begin{IEEEbiography}[{\includegraphics[width=1in,height=1.25in,clip,keepaspectratio]{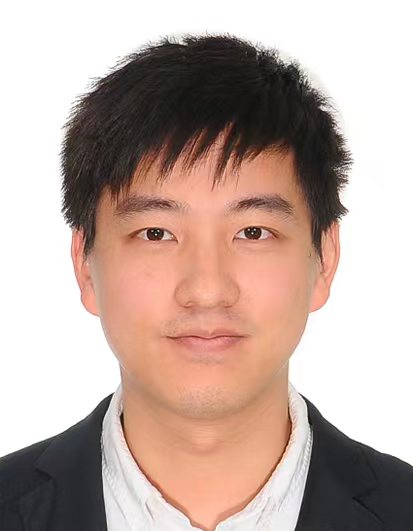}}]{Yicheng Zhang}
	Yicheng Zhang received the Bachelor of Engineering in Automation from Hefei University of Technology in 2011, the Master of Engineering degree in Pattern Recognition and Intelligent Systems from University of Science and Technology of China in 2014, and the PhD degree in Electrical and Electronic Engineering from Nanyang Technological University, Singapore in 2019. He is currently a research scientist at the Institute for Infocomm Research (I2R) in the Agency for Science, Technology and Research, Singapore (A*STAR). Before joining I2R, he was a research associate affiliated with Rolls-Royce @ NTU Corp Lab. He has participated in many industrial and research projects funded by National Research Foundation Singapore, A*STAR, Land Transport Authority, and Civil Aviation Authority of Singapore. He published more than 70 research papers in journals and peer-reviewed conferences. He received the IEEE Intelligent Transportation Systems Society (ITSS) Young Professionals Traveling Scholarship in 2019 during IEEE ITSC, and as a team member, received Singapore Public Sector Transformation Award in 2020.
\end{IEEEbiography}



\begin{IEEEbiography}[{\includegraphics[width=1in,height=1.25in,clip,keepaspectratio]{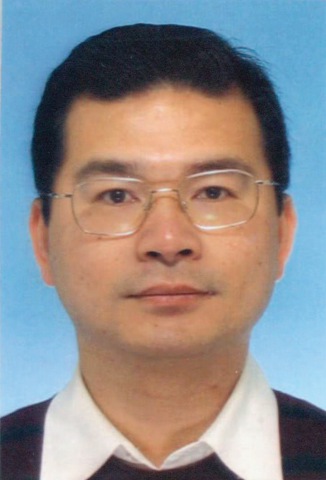}}]{Rong Su}
	received the M.A.Sc. and Ph.D. degrees both	in electrical engineering from the University of Toronto, Toronto, Canada, in 2000 and 2004 respectively.	He is affiliated with the School of Electrical and Electronic Engineering, Nanyang Technological University, Singapore. His research interests include modeling, fault diagnosis and supervisory control of discrete-event dynamic systems. Dr. Su has been a member of IFAC technical committee on discrete event and hybrid systems (TC 1.3) since 2005.
\end{IEEEbiography}

\end{document}